\documentclass[runningheads]{llncs}
\usepackage{graphicx}
\usepackage{cite}
\usepackage{amsmath}
\usepackage[]{algorithm}
\usepackage{algpseudocode}
\algnewcommand\And{\textbf{and}}
\usepackage{amsfonts}
\usepackage{caption}
\usepackage{rotating}
\usepackage{xcolor}
\usepackage{multirow}
\usepackage{color}
\usepackage{array}
\usepackage{booktabs}
\usepackage{tabularx}
\usepackage{multirow}
\usepackage{placeins}
\usepackage{longtable}
\usepackage{supertabular,enumitem}

\usepackage{hyperref}

\usepackage{xspace}

\usepackage{todonotes}

\newcolumntype{L}[1]{>{\raggedright\let\newline\\\arraybackslash\hspace{0pt}}m{#1}}

\newcommand{\prob}{\texttt{EV-CSL}\xspace}
\newcommand{\problemfull}{\texttt{Electric Vehicle Charging Stations Locations}\xspace}

\begin{document}
\title{Citizen centric optimal electric vehicle charging stations locations in a full city: case of Malaga}
%
%
\author{Christian Cintrano\inst{1}\orcidID{0000-0003-2346-2198} \and 
Jamal Toutouh\inst{1,2}\orcidID{0000-0003-1152-0346} \and
Enrique Alba\inst{1}\orcidID{0000-0002-5520-8875}}
\authorrunning{C. Cintrano et al.}
%

\institute{University of Malaga,
 Bulevar Louis Pasteur 35, 29010 Malaga, Spain \\
\email{\{cintrano,jamal,eat\}@lcc.uma.es} \and
Massachusetts Institute of Technology, CSAIL, MA, USA \\
\email{toutouh@mit.edu}
}
\maketitle              
\vspace{-2em}
\begin{abstract}
This article presents the problem of locating electric vehicle (EV) charging stations in a city by defining the \problemfull (\prob) problem. 
The idea is to minimize the distance the citizens have to travel to charge their vehicles. \prob takes into account the maximum number of charging stations to install and the electric power requirements. Two metaheuristics are applied to address the relying optimization problem: a genetic algorithm (GA) and a variable neighborhood search (VNS). The experimental analysis over a realistic scenario of Malaga city, Spain, shows that the metaheuristics are able to find competitive solutions which dramatically improve the actual installation of the stations in Malaga. GA provided statistically the best results.





\vspace{-1em}
\vspace{-2em}
\end{abstract}
%
%
%
%
\section{Introduction}
\vspace{-0.2cm}
Road transportation is one of the main sources of air pollutants in our cities~\cite{lebrusan2020sc}. 
Reducing the road vehicles' emissions would have an important impact on tackling global warming~\cite{Paraschiv2019} and improving inhabitants' health~\cite{lebrusan2020car}. 
An extended use of electric vehicle (EV) transportation will reduce the emission of pollutants. 
However, at the time of this research, the EV adoption is limited by several factors. 
One of the main factors is the need for a specific charging infrastructure for EV, which present two main issues~\cite{Haustein2018}:
\textit{a)} charging times determine the number of vehicles that can be charged over the time; and \textit{b)} charging stations have high energy consumption requirements, which limits the number of stations that can be installed in a given area. 
Improving the availability of charging stations will lead to increasing the adoption of this type of green transportation~\cite{Haustein2018}. 

Smart cities provide a series of tools for advanced knowledge and decision-making support~\cite{camero2019,CINTRANO2020113684,fabbiani2018,lebrusan2020sc}.
In this line, this work focuses on providing a solution to efficiently
allocate EV charging stations to ease EV uptake by the citizens. 
The proposed approach takes into account the distance the users have to travel to charge their EV and the power electric substations installed in the city 
to provide electric power to the different urban areas. These electric substations 
produce a limited energy, which determines the maximum EV charging stations to be allocated in a given area.
Thus, we have defined an optimization problem named \problemfull (\prob). 

Finding the best locations of EV charging stations is attracting the attention of the research community. 
In Brandstätter et al.~\cite{BRANDSTATTER201717} the authors presented an ad-hoc heuristic for solving the \prob problem. The main drawback of their approach is that it does not take into account the energy constraints of the substations. Other research studies only considered the aspects related to the installation of the EV charging points~\cite{HUANG2020102179}, i.e., installation price, maintenance, etc., leaving quality of service (QoS) and users aside of the problem.

The problem of finding the optimal locations for the EV charging stations for a full city can be defined as a variant of a \textit{p}-median problem~\cite{Dantrakul2014a,Megiddot1984}, which have been proven to be an NP-Hard optimization problem. 
The large search space makes impractical the use of traditional optimization methods (e.g., enumeration techniques or dynamic programming). Thus, heuristic and metaheuristics are useful methods to perform the search using bounded computational resources~\cite{COLMENAR201888}. 
Here, two metaheuristic algorithms have been applied to address \prob: a genetic algorithm (GA)~\cite{GA} and a variable neighborhood search (VNS)~\cite{VNS}. 

To evaluate the proposed approach, a realistic scenario of the whole city of Malaga (Spain) has been modeled by taking into account real data (i.e, open data provided by the municipality of the city, road maps from Open Street Maps~\cite{OpenStreetMap}, and electric substations locations).
In turn, the computed results are compared against the current solution provided by the municipality (actual locations of the charging stations), as a baseline solution.


The main contributions of our work are:
\begin{itemize}[noitemsep,topsep=1pt]
    \item Providing the mathematical formulation of \prob.
    \item Modeling a realistic instance based on real data of the city of Malaga to address \prob focusing on citizens' needs and electricity supply constraints.
    \item Implementing and applying two metaheuristics to \prob: a GA and a VNS. 
    \item Studying the solutions computed by the proposed algorithms to analyze their performance when addressing this problem and comparing them with the actual solution deployed in Malaga.
\end{itemize}

The rest of this article is organised as follows: 
the following section defines the \prob addressed in this research.  
Section~\ref{sec:methodology} introduces the main aspects of the metaheuristics applied and implemented. 
Section~\ref{sec:exp-settings} describes the real-world scenario defined to tackle the \prob problem and the main experimental settings. 
The experimental analysis is reported in Section~\ref{sec:exp-analysis}. 
Finally, Section~\ref{sec:conclusion} presents the conclusions and formulates the main lines for future work.

\section{Efficient full city EV charging station locations}
\label{sec:definition}

The \problemfull (\prob) problem is defined to provide potential locations of EV charging stations for a full city given a maximum number of charging stations ($Ms$) to provide the best QoS possible. In this study, as a preliminary approach, the QoS is given by the distance between the citizens' homes and the EV charging station.  
The mathematical formulation of \prob is defined considering the following: 

\begin{itemize}[topsep=1pt]
\addtolength{\itemsep}{2pt}
\item A maximum number of electric car charging station defined by $Ms$.
\item A set $S = \{s_1, \ldots, s_M\}$ of potential street segments for charging stations. For this version of the problem, each street segment of $S$ can be the location of one charging station. 
\item A set $C = \{c_1, \ldots, c_N\}$ of client locations. Following a usual approach in the related literature, nearby locations are grouped in clusters, assuming a similar behavior between elements in each cluster. The number of users to serve at each location $c$ is $u_{c}$. The distance from client $c$ to the charging station $s \in S$ is $dc_{c,s}$, and the maximum distance between any client in $C$ and its assigned charging station (in meters) is $Dc$.
\item A set $E = \{e_1, \ldots, e_T\}$ of electrical substations that serve as electric power source for the charging stations. The distance from the electrical substation $e$ to the charging station in $s \in S$ is $de_{e,s}$, and the maximum distance between substation $e$ in $E$ and its assigned charging station $s$ (in meters) is $De$. 
As the substations have electric power restrictions, the number of charging stations that can be fed by a substation $e$ is limited by $mp_{e}$. 
\end{itemize}
%
Eq.~\ref{eq:1}-\ref{eq:7} describe the model, using the following variables: 
$x_{c,s}$ is 1 if the client $c$ is assigned to the station located in $s$ and 0 otherwise, and
$y_{e,s}$ is 1 if the electrical substation $e$ is feeding the charging station located in $s$ and 0 otherwise.  


\vspace{-0.5cm}
\begin{align}
\small
\label{eq:1}
\min & \sum_
{c\in C, \text{ } s\in S}
x_{c,s}  d_{c,s}  u_{c} \\
\nonumber
\text{subject to} \\
\label{eq:2}
\sum_{s\in{}S} x_{c,s}  ={} & 1 & \forall{}\ c\in{}C & \\
\label{eq:3}
\sum_{c\in C, \text{ } s\in S} x_{c,s}  = & \left\vert{}C\right\vert{} \\
\label{eq:4}
dc_{c,s} x_{c,s}\leq{}&Dc & \forall{}\ c\in{}C,\ s\in{}S & \\
\label{eq:5}
de_{e,s} y_{e,s}\leq{}&De & \forall{}\ e\in{}E,\ s\in{}S & \\
\label{eq:6}
\sum_{s\in{}S} y_{e,s} \leq{} & mp_{e,s} & \forall{}\ e\in{}E & \\
\label{eq:7}
\left\vert{}S_o\right\vert{} ={} & Ms & S_o = \{s_o \setminus \forall{} s_o\in{} S , \sum_{c\in C} x_{c,s_o}  > 0  \} 
\end{align}
\vspace{-0.5cm}

As the QoS provided is measured in terms of distance between the clients and the charging stations, a single objective is provided in Eq.~\ref{eq:1}: minimizing the distance between the clients and the assigned charging stations. 
Regarding the problem constraints,
all the clients are assigned to a unique charging station (Eq.~\ref{eq:2});
all the clients are served for any charging station, i.e., there are not potential clients without a charging station assigned (Eq.~\ref{eq:3});
the maximum distance between the charging station and the client assigned should be lower than $Dc$ (Eq.~\ref{eq:4}); 
the maximum distance between the electric substation and the fed charging station should be lower $De$ (Eq.~\ref{eq:5});  
the number of charging stations that are fed by a given electric substation should be lower or equal than $pm_e$ (Eq.~\ref{eq:6});
and the number of charging stations located should be lower than the maximum number of charging stations to be located $Ms$.




\setlength{\intextsep}{2pt}

\section{Metaheuristics for efficient \prob}
\label{sec:methodology}

This section summarizes the applied metaheuristics to address \prob in the city of Malaga and introduces the main implementation details. 

\subsection{Algorithms}
\label{sec:algorithms}

\subsubsection{Genetic algorithm:} It was originally presented by John Holland inspired by the evolution of species in Nature~\cite{GA}. 
A basic pseudocode is showed in Algorithm~\ref{algorithm:ga}. GA is an iterative method. 
In each iteration, the algorithm generates $\lambda$ new solutions (new population). A new solution is generated from several parent solutions (two parents this case, $p_1$ and $p_2$) selected from the previous population. The selected solutions are mixed (crossover) between them to generate a new one, which is probabilistically disturbed (mutated). At the end of an iteration, the new solutions replace others from the previous population following some kind of strategy. Finally, the algorithm returns the best solution found.

\subsubsection{Variable Neighbourhood Search:} It is based on the concept of neighbourhood~\cite{VNS}. The pseudocode is showed in Algorithm~\ref{algorithm:vns}. Each solution has a defined neighbourhood, i.e., a set of solutions with closest facilities to it. The current solution $x$ is modified according to these neighbourhoods ($next()$ indicates the number of modifications) and improved by a local search. In this version, a number $K$ of consecutive non-improvements is allowed before finishing the algorithm.
The VNS applied in the proposed approach is based on the version defined in~\cite{Drezner2015}.


\begin{minipage}[t]{0.46\textwidth}
\begin{algorithm}[H]
\caption{GA}
\label{algorithm:ga}
\footnotesize
\begin{algorithmic}[1]
\State $pop \gets generatepopulaion()$
\State $i \gets 1$
\While{non stop condition}
    \State $pop ^\prime \gets \emptyset$
    \For{$l \in \{1 ..\lambda\}$}
    	\State $p_1, p_2 \gets select(pop)$
    	\State $x \gets crossover(p_1, p_2)$
    	\State $x^\prime \gets mutation(x)$ 
    	\State $pop^\prime \gets x^\prime$
	\EndFor
	\State $pop \gets replacement(pop, pop^\prime)$
	\State $i \gets i + 1$
\EndWhile
\State \Return{pop}
\end{algorithmic}
\end{algorithm}
\end{minipage}
\hfill
\begin{minipage}[t]{0.46\textwidth}
  \begin{algorithm}[H]
\caption{VNS}
\label{algorithm:vns}
\footnotesize
\begin{algorithmic}[1]
\State $x \gets generation()$
\State $x \gets localsearch(x)$
\State $r \gets true$
\While{$r$ $\And$ \ non stop condition}
	\State $r \gets false$
	\State $j \gets 1$
	\While{$\neg r$ $\And$ $\ j \leq K$}
	\State $i \gets 1$
		\While{$\neg r$ $\And$ $\ i \leq k_{max}$}
			\State $k \gets next(i, k_{max})$
			\State $x^\prime \gets shake(x, k)$
			\State $x^\prime \gets localsearch2(x^\prime)$
			\State $x \gets acceptation(x, x^\prime)$
			\If{$x = x^\prime$} $\ r \gets true$ \Else $\ r \gets false$ \EndIf
			\State $i \gets i + 1$
		\EndWhile
		\State $j \gets j + 1$
	\EndWhile
\EndWhile
\State \Return{x}
\end{algorithmic}
\end{algorithm}
\end{minipage}

\subsection{Implementation details}
\label{sec:implementation}
The solution encoding and the fitness function evaluation are defined in this section. Other details about the applied algorithms (i.e., GA and VNS variation operators) are presented in the parameter settings section (see Section~\ref{sec:exp-analysis}).
\vspace{-0.2cm}
\subsubsection{Solution encoding:} 
The applied solution encoding considers a vector $S^o$ of $S_M$ binary elements, i.e., $S^o = \langle s^o_0, ..., s^o_{S_M}\rangle$, being $S_M$ the number of road segments that are potential locations for the EV charging stations (in the modeled scenario $S_M=33,550$, see Section~\ref{sec:scenario}). Thus, if in the road segment $i$ there is an EV charging station $s^o_i$=1, otherwise $s^o_i$=0. 
\vspace{-0.2cm}
\subsubsection{Fitness function:} 
The fitness function evaluates the QoS provided by installing EV charging stations in the locations represent by the solution $S^o$. In this approach, the QoS is given by the distance that the users have to travel from their homes to the charging station. Thus, the \prob problem is defined as a minimization problem in which the objective function is the average distance the citizens travel from their homes (in \prob are known as neighborhood centers that groups a set of buildings) to the EV charging station. The objective function is defied according to Eq.~\ref{eq:1}.

\vspace{-0.2cm}
\section{Experimental settings}
\label{sec:exp-settings}
\vspace{-0.1cm}

This section describes the main aspects of the experiments carried out to address \prob by using GAs and VNSs. It presents the real-world scenario/instance defined to evaluate the proposed approach. It summarizes the implementation and computational platform. It describes the experiments performed by using irace to configure the main parameters of the applied algorithms.  

\subsection{Scenarios}
\label{sec:scenario}
\vspace{-0.1cm}

\begin{figure}[t]
\includegraphics[width=0.9\linewidth]{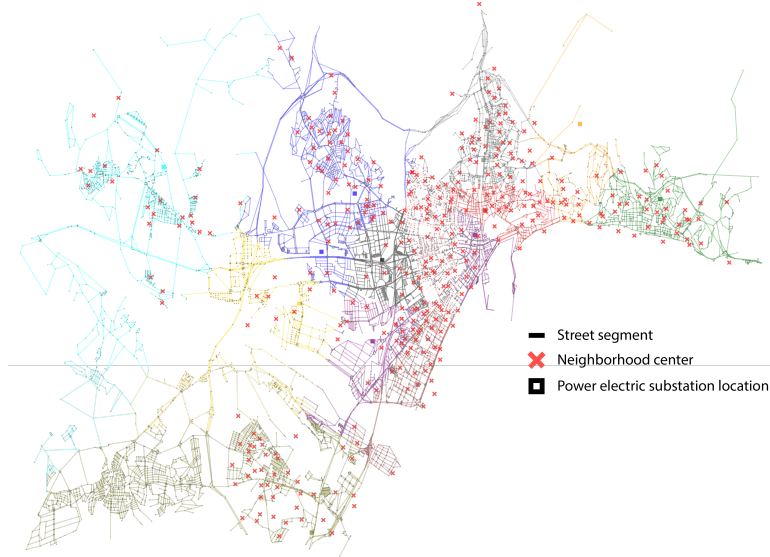}
\vspace{-1em}
\caption{Road map of Malaga, Spain. The edges represent each possible street segment associated with a substation.}
\label{fig:malaga}
\vspace{-0.5cm}
\end{figure}

The \prob is addressed over the city of Malaga, as case study. 
This realistic scenario consists of 567,953 citizens in 363 neighborhoods. 
The road map is defied by using the data of Open Street Maps Based. The map includes total of 33,550 road segments as tentative locations for the charging stations (i.e., $S_M$=33,550). 
Finally, the scenario includes the main data of the actual 14~electrical substations, i.e., locations, maximum energy flow capacity, etc. (see Fig.~\ref{fig:malaga}).
The maximum energy flow capacity limits the number of charging stations that can be located in a given area. 
Thus, it is not realistic to place as many charging stations as we want in any place because a fast charge of a medium-class EV station consumes more than a whole building of apartments.
The different colors in Fig.~\ref{fig:malaga} show the areas of the city covered by each substation.


Different instances have been defined by changing the maximum number of EV charging stations to be installed, i.e, $M_s$. To compare among the different methods, five instances are use with $M_s \in \{10, 20, 30, 40, 50\}$. 
Besides, to compare the provided solutions against the actual one provided by the municipality of Malaga (baseline), another instance was defined with $M_s$=45 because at the time of this research Malaga has 45 EV charging stations. 

\subsection{Implementation and hardware platform}
\label{sec:hw}
The computation platform used in this work consists of a cluster of 144 cores, equipped with three Intel Xeon CPU (E5-2670 v3) at 2.30$\,$GHz and 64$\,$GB memory.
We have carried out 30 runs of each experiment. 
The stop condition for both algorithms is the computational time, in this case, they run for 60 CPU seconds. After that, the algorithms report the best solution found in each of the runs. The algorithms were implemented by using C\texttt{++}, the source code can be found in \url{https://github.com/NEO-Research-Group/EV-CSL}. 

\vspace{-0.2cm}
\subsection{Parameter settings}
GAs and VNSs can use parameters and operators: crossover, mutation, local searches, etc. We have implemented several alternatives for these operators. To get the best parameter setting of the algorithms for our problem, a preliminary parameter setting study was performed. Using a reduced scenario, i.e., the northwest area of Malaga, and irace~\cite{irace} tool to obtain the best configuration of our algorithms.
The two best configurations returned by irace are the ones used in the experimental analysis, trying to avoid possible overfitting in the process carried out by irace. 
Table~\ref{table:tunning-final} shows the configurations of the GA and the VNS.

\begin{table}[t]
\footnotesize
\centering
\setlength\tabcolsep{3pt} 
\caption{Two best parameter configurations found by irace for GA and VNS.}{
\begin{tabular}{lrrclrr}
  \toprule
Parameter	&	GA-1	&	GA-2	&\phantom{ab}&	Parameter	&	VNS-1	&	VNS-2	\\
\midrule											
population	&	30	&	50	&&	Neighbour. model	&	\texttt{Quadr.}~\cite{CINTRANO2020113684}&	\texttt{Closest}~\cite{CINTRANO2020113684}\\		
$\lambda$	&	12	&	6	&&	Neighbour. Size	&	17	&	6	 \\
selection	&	\texttt{Worse one}	&	\texttt{Better one}	&&	shake	&	\texttt{Random}	&	\texttt{Random}	\\
crossover	&	\texttt{None}	&	\texttt{CUPCAP}~\cite{COLMENAR201888}	&&	next	&	\texttt{None}	&	\texttt{None}	\\
mutation	&	\texttt{Random}	&	\texttt{Random}	&&	localsearch	&	\texttt{None}	&	\texttt{IALT}$_{L=20}$~\cite{Drezner2016}	\\
mut. prob.	&	0.65	&	0.76	&&	localsearch2	&	\texttt{FI}~\cite{doi:10.1080/03155986.1983.11731889}	&	\texttt{FI}~\cite{doi:10.1080/03155986.1983.11731889}	\\
replacement	&	$(\mu,\lambda)$	&	$(\mu+\lambda)$ 	&&	k$_{max}$	&	44	&	34	\\
	&		&		&&	K	&	85	&	59	\\
	&		&		&&	accept	&	\texttt{Elitist}	&	\texttt{Elitist}	\\

  \bottomrule
\end{tabular}}
\label{table:tunning-final}
\vspace{-0.5cm}
\end{table}

\vspace{-0.1cm}
\section{Experimental analysis}
\label{sec:exp-analysis}
\vspace{-0.1cm}
This section presents the main results of the experiments carried out by performing 30 independent runs of each algorithm  variation and each of the instances (i.e., $M_s \in \{10,20,30,40,45,50\}$).

\vspace{-0.1cm}
\subsection{Optimization results comparison}

\begin{figure}[!b]
\vspace{-1em}
\includegraphics[width=\linewidth]{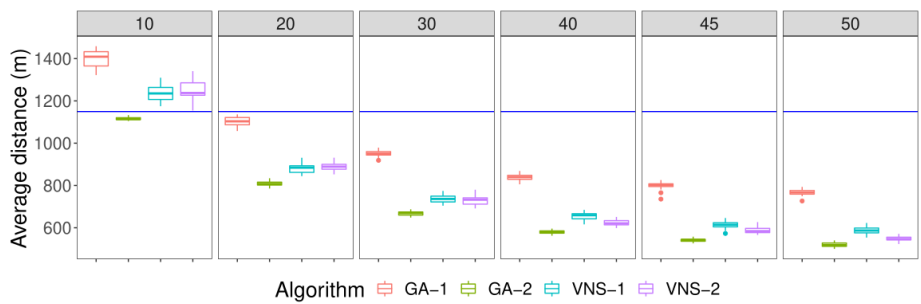}
\vspace{-2em}
\caption{Fitness value (average distance) of each algorithm for the different instances. The blue line represents the fitness value of baseline solution. }
\label{fig:box}
\end{figure}

Fig.~\ref{fig:box} and Table~\ref{tab:statistics} presents the results of each algorithm for the six instances addressed in terms of fitness value (i.e., average distance that users have to travel to get the assigned charging station) of the best solution found. 
The blue line represents the fitness value obtained for the the solution that represents the actual location of the EV charging stations in Malaga (baseline solution). 
Comparing the four metaheuristic alternatives, GA-2 provides the best (lowest) results for all the instances. In turn, GA-2 provides the most robust method because it shows the lowest variability among the different runs. 

According to Wilcoxon Signed Ranks with Bonferroni correction, 
GA-2 is the best method and GA-1 provides the worst results, for all the instances. This remarks the importance of finding the proper configuration of the GA. 
For instances $M_s$=10 and $M_s$=20, VNS-1 and VNS-2 do not show statistical difference. VNS-2 provides statistically the second best results the rest of instances.

As it can be seen in Fig.~\ref{fig:box}, all the proposed algorithms improve the baseline QoS (distance) when installing only 20~stations. 
In turn, GA-2 is able to improve the baseline using only 10~stations.

\begin{table}[t]
\centering
\caption{Experimental results for each algorithm in each instance ($\times 10^2$).}
\label{tab:statistics}
\setlength{\tabcolsep}{4pt}
\renewcommand{\arraystretch}{0.85}
\begin{tabular}{rlrrrr}
  \toprule
$M_s$	&	Algorithm	&	Mean$\pm$SD					&	Min	&	Median	&	Max	\\
\midrule
10	&	GA-1	&	14.08$\pm$0.38	&	13.25	&	14.14	&	14.63	\\
10	&	GA-2	&	\textbf{11.21$\pm$0.075}	&	\textbf{11.03}	&	\textbf{11.12}	&	\textbf{11.31}	\\
10	&	VNS-1	&	12.44$\pm$0.40	&	11.73	&	12.42	&	13.14	\\
10	&	VNS-2	&	12.52$\pm$0.46	&	11.54	&	12.41	&	13.43	\\
\hline
20	&	GA-1	&	11.03$\pm$0.20	&	10.60	&	11.06	&	11.42	\\
20	&	GA-2	&	\textbf{8.08$\pm$0.11}	&	\textbf{7.86}	&	\textbf{8.07}	&	\textbf{8.34}	\\
20	&	VNS-1	&	8.80$\pm$0.22	&	8.44	&	8.84	&	9.31	\\
20	&	VNS-2	&	8.89$\pm$0.20	&	8.52	&	8.89	&	9.31	\\
\hline
30	&	GA-1	&	9.51$\pm$0.14	&	9.19	&	9.50	&	9.79	\\
30	&	GA-2	&	\textbf{6.68$\pm$0.11}	&	\textbf{6.48}	&	\textbf{6.70}	&	\textbf{6.87}	\\
30	&	VNS-1	&	7.36$\pm$0.19	&	7.04	&	7.36	&	7.74	\\
30	&	VNS-2	&	7.28$\pm$0.21	&	6.92	&	7.33	&	7.80	\\
\hline
40	&	GA-1	&	8.40$\pm$0.15	&	8.06	&	8.40	&	8.69	\\
40	&	GA-2	&	\textbf{5.80$\pm$0.08}	&	\textbf{5.62}	&	\textbf{5.81}	&	\textbf{5.96}	\\
40	&	VNS-1	&	6.55$\pm$0.16	&	6.16	&	6.59	&	6.84	\\
40	&	VNS-2	&	6.23$\pm$0.14	&	5.98	&	6.21	&	6.51	\\
\hline
45	&	GA-1	&	7.99$\pm$0.17	&	7.35	&	8.02	&	8.26	\\
45	&	GA-2	&	\textbf{5.42$\pm$0.08}	&	\textbf{5.26}	&	\textbf{5.43}	&	\textbf{5.58}	\\
45	&	VNS-1	&	6.12$\pm$0.16	&	5.73	&	6.14	&	6.46	\\
45	&	VNS-2	&	5.88$\pm$0.16	&	5.66	&	5.84	&	6.27	\\
\hline
50	&	GA-1	&	7.66$\pm$0.13	&	7.27	&	7.67	&	7.93	\\
50	&	GA-2	&	\textbf{5.20$\pm$0.10}	&	\textbf{5.00}	&	\textbf{5.18}	&	\textbf{5.40}	\\
50	&	VNS-1	&	5.86$\pm$0.16	&	5.54	&	5.87	&	6.23	\\
50	&	VNS-2	&	5.49$\pm$0.10	&	5.23	&	5.47	&	5.71	\\

   \bottomrule
\end{tabular}
\vspace{-2em}
\end{table}

\subsection{Improvement on travel distance over the real layout of stations}

To better illustrate the improvement offered by our algorithms versus the actual stations locations in the city of Malaga (a.k.a. baseline solution), we have compare the solutions found by installing the same number of stations ($M_s$=45) in terms of average distance that the EV users have to travel to charge their cars.

\begin{figure*}[t]
\includegraphics[width=\linewidth]{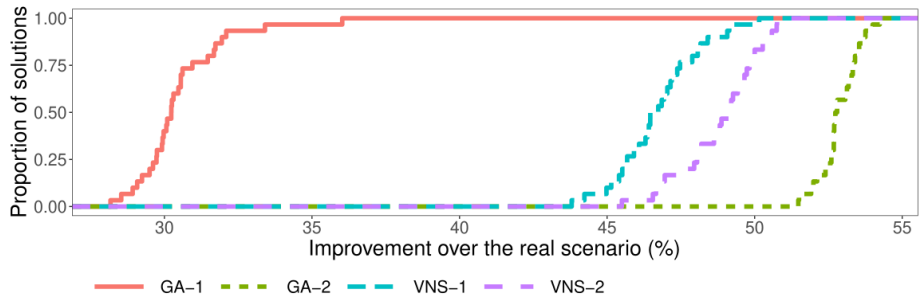}
\vspace{-2em}
\caption{Empirical cumulative distribution of the percentage of improvement of our solutions in each algorithm,
compared to the baseline solution.}
\label{fig:fitness-comp-algo}
\vspace{-1em}
\end{figure*}

Fig.~\ref{fig:fitness-comp-algo} shows the proportion of solutions for each algorithm (\textit{y}-axis) obtained less than the percentage of improvement defined in the \textit{x}-axis, i.e., the percentage of computed solutions that achieved at most that percentage of improvement.
GA-1 lags far behind the other algorithms by only over 30\% in the 75\% of its solutions. The two versions of VNS offer improvements between 40-50\%. The best algorithm is GA-2, being also the most stable (less steep curve) with a 52\% of improvement in more than the 70\% of its solutions. In general, it is interesting to note that the algorithms using the second-best configurations found by irace offer the most significant improvements. This result underlines the importance of take into account the overfitting when we configuring machine learning techniques.

\section{Conclusion}
\label{sec:conclusion}
This article presented a definition of the \prob optimization problem.
The optimization problem takes into account the QoS provided (in terms of distance customers have to travel to get the charging station) and the energy limitations of the different electric substations around the city.
Two different metaheuristic algorithms, both parameterized using irace, have been proposed and implemented to address the problem: GA and VNS. 

A realistic scenario based on city of Malaga has been defined by using real data (i.e., road maps, inhabitants' home location, electric substations location, etc.) 
Different instances have been defined by locating a different number of charging stations (from 10 to 50). 

The main results of the experimental evaluation indicate that the proposed metaheuristics were able find competitive solutions. The solutions provided by the proposed methodology were able to improve the actual QoS provided in Malaga with 45~stations installing only 20. 
In general, a variation of GA provided the best results for the different instances. 
When comparing the actual solution in the city with the ones provided by the four metaheuristic variations analyzed here, metaheuristic dramatically improve the QoS.

The main lines for future work are related to extending the proposed problem model to consider the number of parking slots in each station and the charging time, exploring other optimization methods, and
defining a multi-objective variation of the problem by including other objectives, such as the installation costs.
In addition, we are working to improve the proposed model to include in the QoS the idea that the vehicles can be charged when the citizens are working or doing other activities. 



\vspace{-1em}
\section*{Acknowledgment}
This research was partially funded by the University of M\'alaga, Andaluc\'{\i}a Tech, the Junta de Andaluc\'ia UMA18-FEDERJA-003 and the project TAILOR Grant \#952215, H2020-ICT-2019-3.
J. Toutouh research was partially funded by European Union’s Horizon 2020 research
and innovation program under the Marie Skłodowska-Curie grant agreement
No 799078.

\bibliographystyle{splncs04}
\bibliography{main}

\end{document}